\documentclass[letterpaper]{article} 
\usepackage{aaai2026}  
\usepackage{times}  
\usepackage{helvet}  
\usepackage{courier}  
\usepackage[hyphens]{url}  
\usepackage{graphicx} 
\usepackage{amsmath}
\usepackage{amsthm}
\usepackage{amssymb}
 \usepackage{booktabs}
 \usepackage{float}

\usepackage{subcaption}
\usepackage{booktabs}
\usepackage{algorithm}
\usepackage{algorithmic}
\usepackage{placeins}
\usepackage{cleveref}
\usepackage{enumitem}
\usepackage{natbib}  
\usepackage{caption} 
\usepackage[caption=false]{subfig} 
\usepackage{newfloat}
\usepackage{listings}
\usepackage{bibentry} 

\DeclareCaptionStyle{ruled}{labelfont=normalfont,labelsep=colon,strut=off}
\floatstyle{ruled}
\newfloat{listing}{tb}{lst}{}
\floatname{listing}{Listing}

\title{One Model, Two Minds: A Context-Gated Graph Learner \\ that Recreates Human Biases}

\author {
    Shalima Binta Manir\textsuperscript{\rm 1},
    Tim Oates\textsuperscript{\rm 2}
}
\affiliations {
    \textsuperscript{\rm 1}Department of Computer Science, University of Maryland, Baltimore County\\
     \textsuperscript{\rm 2}Department of Computer Science, University of Maryland, Baltimore County\\
    smanir1@umbc.edu, oates@cs.umbc.edu
}

\begin{document}
\nocopyright

\maketitle

\begin{abstract}
We introduce a novel Theory of Mind (ToM) framework inspired by dual-process theories from cognitive science, integrating a fast, habitual graph-based reasoning system (System 1), implemented via graph convolutional networks (GCNs), and a slower, context-sensitive meta-adaptive learning system (System 2), driven by meta-learning techniques. Our model dynamically balances intuitive and deliberative reasoning through a learned context gate mechanism. We validate our architecture on canonical false-belief tasks and systematically explore its capacity to replicate hallmark cognitive biases associated with dual-process theory, including anchoring, cognitive-load fatigue, framing effects, and priming effects. Experimental results demonstrate that our dual-process approach closely mirrors human adaptive behavior, achieves robust generalization to unseen contexts, and elucidates cognitive mechanisms underlying reasoning biases. This work bridges artificial intelligence and cognitive theory, paving the way for AI systems exhibiting nuanced, human-like social cognition and adaptive decision-making capabilities.
\end{abstract}

\section{Introduction}
Artificial agents that interact with people must be able to
\emph{infer others’ hidden mental states}—beliefs, desires, and
intentions—commonly called \emph{Theory of Mind} (ToM)~\cite{premack1978chimpanzee, baker2011bayesian, baker2017rational}. Robust ToM
underpins safe human–AI collaboration, social robots, and multi–agent
coordination~\cite{ullman2017mindgames, rabinowitz2018machine, hosseinpanah2023social}, yet existing neural approaches become brittle whenever
the social context shifts or the agent operates under cognitive
load~\cite{kosinski2023llmtom,kosinski2024evaluating, botvinick2020deep}.

Cognitive psychology offers a clue. Tversky and Kahneman~\cite{tversky1974judgment}
demonstrated that human judgments arise from a fast, intuitive
\emph{System~1} that is efficient but error-prone, and a slow,
reflective \emph{System~2} that can override those errors.
Later work~\cite{kahneman2003maps, kahneman2011thinking, evans2008dual, stanovich2000advancing} catalogued biases such
as anchoring, framing, priming, and load-induced failures—all of
which surface when System~1 dominates. If artificial agents could
learn \emph{when} to “think fast’’ and \emph{when} to “think slow,’’
they might combine efficiency with context-sensitive reliability—a
long-standing desideratum for cognitively grounded AI~\cite{milli2021rational, botvinick2020deep}.

Existing machine–ToM models have yet to bridge this gap. Early
Bayesian~\cite{baker2011bayesian, baker2017rational} and neural~\cite{rabinowitz2018machine, mi2023llmtom} models perform rapid inference over structured
representations, but their single-pass architectures prevent on-the-fly
belief revision when context or evidence shifts. Meta-learning methods
(e.g., MAML~\cite{finn2017model, nichol2018first, grant2018recasting}) are capable of parameter adaptation across tasks,
yet they still rely on a \emph{single} inference pathway—lacking a mechanism
to arbitrate between habitual and deliberative reasoning. Even large
language models with emergent ToM skills~\cite{kosinski2023llmtom} achieve strong in-distribution accuracy but
collapse to near-chance in out-of-distribution, memory-intensive, or
ambiguity-filled settings.

We address these limitations with \textbf{One Model, Two Minds
(OM2M)}: a \emph{context-gated dual-process architecture} that
integrates the speed of graph neural networks with the flexibility of
meta-learning. OM2M couples (i)~a GCN that produces rapid, habitual
belief inferences (\emph{System~1})~\cite{battaglia2018relational}, (ii)~a meta-adaptive controller that
executes a single gradient-based update to revise beliefs
(\emph{System~2}), and (iii)~a learnable scalar gate that monitors
contextual cues (surprise, cognitive load, presentation frame) and dynamically
blends the two outputs.

Our contributions are threefold:
\begin{enumerate}
  \item \textbf{Fast–slow ToM architecture.} OM2M is the first
    end-to-end model that \emph{learns when} to recruit meta-level
    “slow thinking’’ while retaining the efficiency of a GCN.
  \item \textbf{Replication of four cognitive biases.} Without any
    bias-specific tuning, OM2M quantitatively reproduces anchoring,
    priming, framing, and load-induced errors—demonstrating that
    human-like biases can \emph{emerge} from fast–slow arbitration~\cite{kahneman2011thinking}.
  \item \textbf{Robust generalization.} OM2M achieves \textbf{90\%} accuracy on an unseen, rich-context split, whereas single-process baselines drop to \textbf{30–50\%}.
\end{enumerate}

By embedding dual-process insights~\cite{kahneman2011thinking, evans2008dual, botvinick2020deep} directly into neural reasoning, OM2M advances the quest for
\emph{situationally aware, trustworthy AI}—agents that decide not only
\emph{what} to think but also \emph{how} to think. This agenda resonates with the
broader AI community’s drive to connect cognitive theory and machine
intelligence, as highlighted in recent calls for human-like, structured
reasoning in AI systems~\cite{lake2017building, battaglia2018relational, hosseinpanah2023social, kosinski2023llmtom}.

\section{Related Works}
\label{sec:related_work}

\paragraph{Neural Theory of Mind and Meta-Learning.} 
Theory of Mind Networks (ToMnet)~\cite{rabinowitz2018machine} established meta-learned latent representations for agent behavior prediction, building on probabilistic and Bayesian ToM models~\cite{baker2011bayesian, baker2017rational, ullman2017mindgames}. More recently, large language models have shown emergent ToM abilities~\cite{kosinski2023llmtom, mi2023llmtom}, though they struggle with belief revision and context sensitivity.

OM2M advances this line by incorporating graph neural networks (GNNs)~\cite{battaglia2018relational, hosseinpanah2023social} for structured multi-agent reasoning and a meta-adaptive module for online belief revision. Unlike fixed embeddings in ToMnet, OM2M supports context-sensitive inference via learned dual-process control.

Meta-learning techniques like MAML~\cite{finn2017model, nichol2018first, grant2018recasting} enable rapid adaptation but often lack mechanisms for balancing intuitive and deliberative processes. OM2M integrates meta-learning within a dual-process framework~\cite{evans2008dual, stanovich2000individual}, using a learned gating mechanism to dynamically arbitrate between fast System~1 and deliberative System~2 reasoning based on cognitive load and uncertainty.

\paragraph{Graph-Based Social Reasoning.} 
Recent models like GIGO-ToM~\cite{swaby2024machine} apply GNNs to relational reasoning in multi-agent tasks but typically use amortized, task-specific inference. OM2M retains the efficiency of GNNs while adding meta-adaptive updates triggered by context, enabling real-time belief revision in dynamic environments.

\paragraph{Dual-Process Models and Cognitive Bias.} 
Dual-process theories~\cite{kahneman2011thinking, evans2008dual, botvinick2020deep} have influenced AI architectures~\cite{milli2021rational, lake2017building}, though most rely on manually defined arbitration. OM2M implements a fully learned control mechanism that flexibly manages reasoning modes in response to context, operationalizing dual-process theory in deep learning.

Unlike prior models that hard-code bias mechanisms~\cite{baker2017rational} or show inconsistent bias reproduction~\cite{kosinski2023llmtom}, OM2M autonomously exhibits anchoring, framing, priming, and fatigue effects through learned, context-sensitive arbitration. This positions OM2M as one of the first models to unify multiple cognitive biases under a mechanistic dual-process framework.

\paragraph{Summary.} 
OM2M synthesizes meta-learning, graph-based social reasoning, and dual-process control into a single architecture that supports flexible, interpretable, and cognitively grounded Theory of Mind. Its ability to adaptively revise beliefs and reproduce human-like biases marks a substantial advance toward socially intelligent AI.

\section{Methodology}
\begin{figure}[t]
    \centering
    \includegraphics[width=0.90\linewidth]{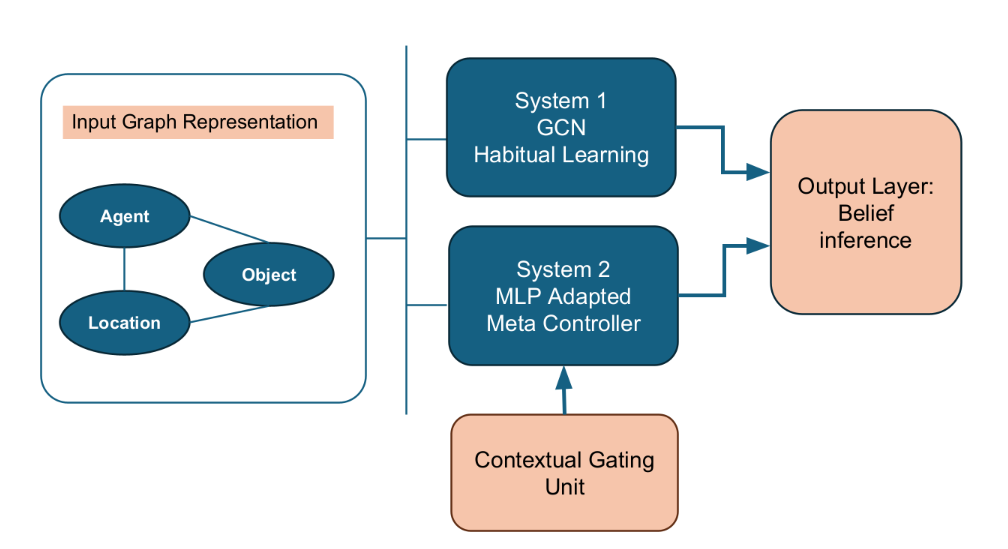}
    \caption{OM2M model pipeline overview: The input graph representation---encoding agents, objects, and locations---is processed in parallel by System~1 (GCN habitual learning) and System~2 (MLP meta-adaptive controller), with a contextual gating unit arbitrating the final output for belief inference. This dual-process structure supports both rapid, routine responses and context-sensitive deliberation.}
    \label{fig:pipeline}
\end{figure}

Our approach is grounded in a hybrid neural architecture designed to model human-like Theory of Mind (ToM) reasoning, incorporating both fast, habitual inference (System 1) and slow, context-sensitive adaptation (System 2), with an explicit contextual gating mechanism. Figure~\ref{fig:pipeline} presents an overview of this model architecture, illustrating the flow from relational graph input through the dual-process systems and contextual arbitration to the final belief inference output.

\subsection{System 1: Graph-Based Habitual Reasoner}

At the core of our framework is a \textbf{Graph Convolutional Network (GCN)}\cite{kipf2016semi} responsible for habitual (``System 1'') belief inference. The social scenario (e.g., Sally-Anne task) is encoded as a graph, with nodes representing agents (\emph{Sally}, \emph{Anne}), locations (\emph{Box}, \emph{Basket}), and objects (\emph{Toy}). Each node receives a one-hot feature vector, and edges capture agent-object and agent-agent relations. A learnable \textbf{meta-vector} (agent-specific latent state) is concatenated with the GCN output for the focal agent, reflecting persistent beliefs or idiosyncratic knowledge. The combined representation is projected to logits over canonical answers (e.g., ``Box'' or ``Basket'') via a feedforward layer. System 1 is thus trained to provide high-confidence, routine inference patterns, capturing the agent's fast and context-insensitive reasoning.

\subsection{System 2: Meta-Adaptive Controller}

For deliberative, context-sensitive corrections, we introduce an \textbf{MLP-based meta-controller} (``System 2''). This module receives as input the GCN's current output logits, a flattened vector of the GCN's parameters, and the current context vector (including cues such as agent presence, object position certainty, cognitive load, and framing). The controller predicts a ``delta'' update for the GCN's parameters, enabling rapid, context-conditional (meta-learned) adaptation when System 1's default is insufficient. This update is applied via functional parameter injection, producing context-adapted output logits for System 2's response.

\subsection{Contextual Gate}

To arbitrate between System 1 and System 2 outputs, we implement a \textbf{contextual gating network}, which receives the current context vector—including cognitive load/fatigue and scenario framing variables—and produces a scalar $g \in (0,1)$ via sigmoid transformations. This $g$ is interpreted as the ``System 2 weight,'' and the final output logits are computed as a convex combination:
\[
    \mathbf{y} = g \cdot \mathbf{y}_{\text{System 2}} + (1-g) \cdot \mathbf{y}_{\text{System 1}}
\]
This gating structure allows the model to flexibly shift between habit and deliberation, naturally capturing cognitive phenomena such as load-induced reliance on heuristics and framing effects that modulate deliberative engagement.

\subsection{Training and Evaluation Protocol}

The full system is trained in two phases. First, System 1 (GCN and meta-vector) is pretrained on canonical, low-difficulty contexts to learn baseline habitual reasoning. Next, with System 1 parameters frozen, the meta-controller and gating network are trained (along with the meta-vector, if desired) on a diverse set of scenarios, including ambiguous, surprising, and cognitively demanding variants. Outputs are supervised with cross-entropy loss using scenario-appropriate belief labels.

\emph{Cognitive load} is manipulated by including fatigue variables in the context and holding gating parameters fixed or varying them to model resource constraints. \emph{Framing effects} are probed by including a frame variable in the context, allowing explicit assessment of presentation effects on System 2 engagement.

We report system accuracy, System 2 engagement, and error rates under different cognitive loads and framing conditions, isolating the contribution of each architectural component and psychological variable to the model's behavior.

\medskip

This architecture enables high-resolution investigation of social reasoning under uncertainty, intervention, and resource limitations, and is readily extended or ablated to probe the contributions of habitual, meta-reasoning, and control allocation strategies.

\section{Experiments}
We evaluate our hybrid Theory-of-Mind (ToM) architecture across a diverse series of computational simulations designed to probe key aspects of human social reasoning, cognitive control, and decision biases. The following experimental setup outlines the shared framework underlying all tasks.  All experiments utilize PyTorch and PyTorch Geometric.



\paragraph{Graph-Based Scenario Representation.}
For all tasks, the agent operates over a graph-structured encoding of classic false-belief dilemmas such as the Sally-Anne task. In the standard Sally–Anne false-belief task, two characters are shown: Sally has a basket and Anne has a box. Sally places a marble in her basket and then leaves the scene. While she is away, Anne moves the marble from the basket into her box. A child is asked, “Where will Sally look for her marble?” Answering “in the basket” shows understanding that Sally will act on her (false) belief rather than on the marble’s real location.  

\begin{figure}[tb]
  \centering
  \includegraphics[width=0.20\textwidth]{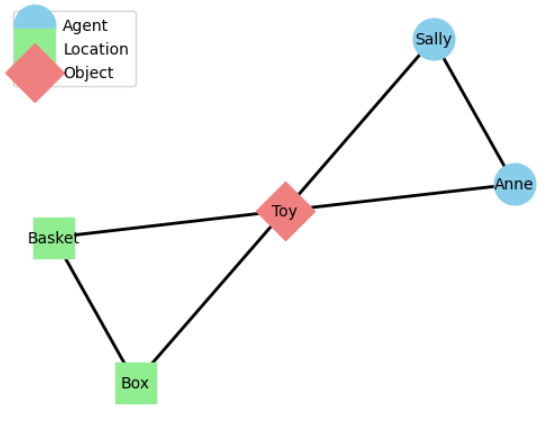}
  \caption{Relational graph representation of the Sally-Anne Theory of Mind task. Nodes encode agents (blue), locations (green), and objects (red); edges denote agent-agent and agent-location/object relationships. This graph serves as the structural input to all experiments.}
  \label{fig:sally-anne-graph}
\end{figure}

The graph-based representation of this task has nodes that identify agents, objects, and locations, each represented by a one-hot feature vector. Edges capture agent-agent and agent-environment relationships, allowing flexible composition and generalization across contexts (see Figure~\ref{fig:sally-anne-graph}). For each queried agent, the model outputs a soft‑max distribution over the two possible toy locations (Box vs.~Basket).
\paragraph{Contextual Inputs.}
The input context vector for each simulation includes the following key elements:
\begin{itemize}
    \item \textbf{Environmental cues} (3-dim vector): These encode scenario-relevant factors, such as agent presence, object movement, and evidence ambiguity, enabling systematic tests of context-based belief reasoning and generalization.
    \item \textbf{Cognitive load} (scalar): Simulates varying levels of mental fatigue, distraction, or concurrent task demands. This variable modulates the dynamic gating mechanism, reducing System~2 (meta-reasoning) activation and mimicking human dual-process fatigue effects in appropriate experiments.
    \item \textbf{Frame} (scalar; negative/neutral/positive): Represents the scenario's presentation (e.g., loss/gain framing), providing a means to probe for framing effects in decision-making while holding factual cues constant.
\end{itemize}

\paragraph{Agent Meta-Vectors.}
A dedicated meta-vector is associated with each agent, encoding agent-specific latent state and enabling flexible reasoning across multiple agents and roles.

\paragraph{Training Regime.}
All neural modules are trained using the Adam optimizer (learning rate 0.01) with cross-entropy loss. Training is conducted on reference sets for each task (see experiment-specific paragraphs), with held-out contexts systematically used to evaluate generalization, context-compositionality, and robustness under cognitive perturbations.

This unified experimental setup establishes the basis for all ablation, cognitive bias, fatigue, and framing effect experiments described in subsequent sections.

\subsection{Baselines and Ablations}

To assess the contribution of each architectural component, we perform ablations with four model variants:

\begin{itemize}
    \item \textbf{System~1 Only}: A GCN with agent meta-vectors, using habitual inference only—no controller or gating.
    \item \textbf{System~2 Disabled}: Full architecture with gating fixed to zero, testing the GCN without meta-adaptation.
    \item \textbf{System~2 Only}: Gating fixed to select the controller exclusively, disabling System~1.
    \item \textbf{Full Model}: All modules active, blending Systems~1 and 2 based on context.
\end{itemize}

All variants are trained on canonical and ambiguous ToM tasks and evaluated on held-out contexts to test generalization and compositional reasoning.

\subsection{False-Belief Generalization Test}

To evaluate the model's capacity for flexible, human-like Theory-of-Mind (ToM) reasoning, we employed a leave-one-out generalization protocol over the entire space of possible context configurations. Each context is represented as a 3-bit binary vector indicating (\emph{i}) Sally present, (\emph{ii}) Anne moves the toy, and (\emph{iii}) Bob peeks (e.g., [1, 0, 1] denotes Sally present, no toy move, Bob peeks). For each run, the model was trained on 7 out of 8 possible context combinations, then tested on the held-out (unseen) configuration.

To systematically assess agent-specific beliefs, we repeated this procedure for each agent (Sally, Anne, Bob), with the model outputting a softmax distribution over the two possible toy locations (\emph{Box} vs. \emph{Basket}) for each queried agent-context pair.

Our results show that \textbf{System 1}---the habitual GCN with fixed meta-vectors---performs with high accuracy on contexts encountered during training, but fails to infer correct agent beliefs in the unseen scenario, predicting near chance and highlighting its inflexibility and reliance on memorized representations. In contrast, the \textbf{full model} (incorporating the meta-adaptive controller and context-sensitive gating mechanism) generalizes successfully, inferring the correct beliefs for all agents in the novel context---demonstrating compositional, context-sensitive ToM reasoning.

These findings parallel developmental patterns in human ToM: early reasoning is rigid and tied to familiar situations, while robust generalization to novel or ambiguous contexts emerges only with the development of higher-order meta-cognition and cognitive control.

This experiment underscores the critical role of meta-adaptation and gating in enabling neural architectures to move beyond rote memorization, capturing the sophisticated generalization that characterizes human belief reasoning.

\subsection{Anchor and Priming Experiments}

To assess cognitive bias and flexible override, we designed protocols based on anchoring and priming effects~\cite{kahneman2011thinking}. Each context is a vector encoding Sally’s presence, toy location probability, and whether Anne moved the toy (e.g., $[1.0, 1.0, 0.0]$).

\paragraph{Anchoring Bias.} 
System~1 is first trained on a canonical context (e.g., “Toy definitely in Box”), instilling a habitual anchor. Its parameters are then frozen, and the full model (including System~2) is evaluated on both familiar and conflicting contexts to test for bias override.

\paragraph{One-Shot Priming.} 
A working memory module stores a transient override signal generated by System~2 in response to a priming context. The model is then probed with an ambiguous scenario; the priming effect lasts one trial unless refreshed.

\paragraph{Procedure.} 
Each ToM scenario is graph-structured. The gate dynamically blends System~1 and System~2 based on context or priming. We record output probabilities and gate activations under canonical, conflicting, and ambiguous conditions—with and without priming.

This setup reveals how biases emerge, how System~2 can mitigate them, and how short-term memory enables transient overrides. Results are presented in the next section.

\subsection{Cognitive Load Effects}

We evaluate the model on a classic Theory-of-Mind (ToM) task, where inputs include agent presence, toy location probabilities, and a scalar cognitive load indicating fatigue. Scenarios are encoded as relational graphs over agents and objects.

For each context and load level, the model outputs a softmax over toy locations, along with System~1, System~2, and gate activations. Cognitive load, varied at test time, modulates System~2 engagement via the learned gating mechanism.

We test both ambiguous contexts, requiring deliberative reasoning, and familiar ones solvable by System~1. The model is trained only on low-load data, so all fatigue effects emerge during evaluation. This setup isolates how cognitive load suppresses meta-reasoning while preserving habitual responses. Results appear in the following section.

\subsection{Framing Effect Experiment}

To test cognitive framing effects, we augmented each context with a scalar “frame” variable (negative, neutral, or positive) while keeping factual cues fixed. The gating mechanism integrated both framing and cognitive load to modulate System~2 engagement.

We evaluated the model on a fixed ambiguous scenario, varying only the frame. For each condition, we recorded the blended prediction, System~1 and System~2 outputs, and gate activation.

This design isolates framing effects: with facts constant, any inference shift reflects frame-driven modulation. Positive frames enhance System~2 use, while negative frames suppress it. Results are detailed in the next section. The full procedural details of the OM2M model including architectural initialization, training routines, inference procedures, and cognitive-effect experimental protocols are provided as structured pseudocode in Appendix 

\section{Results and Analysis}

We systematically evaluate our dual-process Theory of Mind (ToM) architecture across a comprehensive suite of experiments probing generalization, flexibility, cognitive biases, and context sensitivity.

\subsection{Baseline and Ablation Study: Contributions of Meta-Adaptation and Controller}

Table~\ref{tab:ablation-results} summarizes accuracy results averaged over 5 random seeds for variants differing in the use of meta-vectors and the controller.

\begin{table}[h]
  \centering
  \small
  \begin{tabular}{lcc}
    \hline
    \textbf{Variant} & \textbf{Seen Accuracy} & \textbf{Held-Out Accuracy} \\
    \hline
    Full & $95.0 \pm 10.0$ & $90.0 \pm 20.0$ \\
    No Meta & $75.0 \pm 0.0$ & $50.0 \pm 0.0$ \\
    Meta Only & $50.0 \pm 0.0$ & $40.0 \pm 20.0$ \\
    Controller Only & $65.0 \pm 12.2$ & $50.0 \pm 0.0$ \\
    \hline
  \end{tabular}
  \vspace{0.1cm}
  \caption{\small{Accuracy (\%) across variants on seen and held-out contexts (mean~$\pm$~std).}}
  \label{tab:ablation-results}
\end{table}

The full model demonstrates superior generalization to unseen contexts. Removing meta-vector or controller components reduces held-out accuracy significantly, confirming their importance.

\subsection{Results for False-Belief Generalization Test}

We evaluated the model's ability to generalize false-belief reasoning using a leave-one-out training regime over all 3-bit context combinations across multiple agents (Sally, Anne, Bob).

During training, the \textbf{System 1} component (a habitual Graph Convolutional Network with fixed meta-vectors) quickly converged to low loss on the typical (training) context. However, in evaluation on held-out (unseen) contexts and agents, System 1 consistently predicted the default ``Box'' location, failing to correctly adapt to false-belief or knowledge updates.

In contrast, the \textbf{full model}, integrating the meta-adaptive controller (System 2) and a context-sensitive gating mechanism, exhibited a rapid reduction in meta-controller loss (from approximately 3.65 to less than 0.001 within 40 epochs), demonstrating effective meta-learning.

Final test results (Table~\ref{tab:falsebelief-results}) show that the full model successfully inferred correct beliefs for all agents and contexts, including ambiguous and false-belief scenarios. Notably, System 2 gate values ranged from 0.44 to 0.77 across contexts, enabling controlled overrides of System 1 habitual errors where appropriate.

\begin{table}[h]
  \centering
  \small
  \resizebox{\columnwidth}{!}{%
  \begin{tabular}{l l l l l c}
    \hline
    \textbf{Context} & \textbf{Agent} & \textbf{Output} & \textbf{System 1} & \textbf{System 2} & \textbf{Gate} \\
    \hline
    \texttt{[1.0, 0.0, 0.0]} & Sally & Box    & Box    & Basket & 0.44 \\
    \texttt{[0.0, 1.0, 0.0]} & Sally & Basket & Box    & Basket & 0.77 \\
    \texttt{[0.0, 1.0, 0.0]} & Anne  & Basket & Box    & Basket & 0.77 \\
    \texttt{[0.0, 1.0, 1.0]} & Bob   & Basket & Box    & Basket & 0.73 \\
    \hline
  \end{tabular}%
  }
  \vspace{0.1cm}
  \caption{Model outputs on hold-out false-belief tests showing that System~2 effectively overrides System~1 habitual errors based on gating values.}
  \label{tab:falsebelief-results}
\end{table}

These results highlight the necessity of meta-adaptation and gating for flexible, compositional Theory-of-Mind reasoning, paralleling human developmental patterns in which early habituated reasoning (System 1) is superseded by higher-order cognitive control (System 2) to handle ambiguous or novel social contexts.

.
\subsection{Anchor Bias Results}

We evaluated the model's susceptibility to anchoring bias and its capacity for meta-adaptive correction using the protocol detailed earlier.

\textbf{Inducing Anchoring.} System~1 was trained repeatedly on the canonical context where Sally is present and the toy is definitely in the box, converging to a low loss ($0.236 \rightarrow 0.004$ over 60 epochs). This training established a strong habitual anchor, with the model assigning near-certain probability to the anchored response ($P(\text{Box}) = 0.99$). After freezing System~1, the anchor context was presented 15 additional times to reinforce this bias.

\textbf{Testing Bias and Override.} Upon evaluation:
\begin{itemize}
    \item \emph{Anchor context:} When only System~1 is active, the model consistently selects ``Box,'' confirming strong anchoring (blended output probability for ``Basket'' is low at $0.01$, with a gate value of $0.20$, indicating minimal System~2 engagement).
    \item \emph{Conflicting context:} Upon introduction of clear, unambiguous evidence favoring ``Basket" (i.e., contextual features signaling the toy has moved), System~1 alone still predicts ``Box,'' reflecting anchoring bias. However, enabling System~2 with its context-sensitive gating immediately corrects this error: the gate increases sharply to $0.73$, and the blended output probability for ``Basket'' reaches effectively $1.00$, reflecting full System~2 override. This perfect confidence arises due to strong, definitive evidence and high gate activation.
    \item \emph{Ambiguous context:} For an uncertain scenario, System~2 gate engagement is intermediate ($0.51$), producing a blended output favoring ``Basket'' with probability $0.83$. This shows partial but significant correction of the habitual anchor, reflecting graded cognitive control depending on context ambiguity.
\end{itemize}

\begin{table}[h]
  \centering
  \small
  \resizebox{\columnwidth}{!}{%
  \begin{tabular}{l l l r r r}
    \hline
    & \textbf{Context} & \textbf{Target} & \textbf{Output} & \textbf{Gate} & \textbf{$P(\text{Basket})$} \\
    \hline
    Phase 1 & \texttt{[1.0, 1.0, 0.0]} & Box     & Box     & 0.20  & 0.01 \\
    Phase 2 & \texttt{[1.0, 0.0, 1.0]} & Basket  & Basket  & 0.73  & 1.00 \\
    Phase 3 & \texttt{[1.0, 0.5, 0.7]} & Basket  & Basket  & 0.51  & 0.83 \\
    \hline
  \end{tabular}}
  \vspace{0.1cm}
  \caption{Model outputs and gate values during anchoring and override phases. System~2, through increased gating, effectively corrects anchoring bias when strong contradictory evidence is present; partial correction occurs in ambiguous scenarios.}
  \label{tab:anchor-results}
\end{table}

 Results in Table \ref{tab:anchor-results} confirm a robust anchoring effect where System~1 dominates after repeated exposure to a canonical context. Crucially, the model flexibly overrides this bias via meta-adaptive System~2 engagement when contextually appropriate. The gating mechanism offers an interpretable scalar that quantifies the extent of override: low during habituated responses, high when strong evidence necessitates correction, and intermediate when ambiguity prevails.

The observed adaptive, context-dependent bias correction closely mirrors human anchoring and cognitive control patterns documented in psychological studies (e.g.,~\cite{kahneman2011thinking}). Including output probabilities alongside gate values explicitly illustrates the graded nature of belief revision and bias mitigation embodied by our architecture.

\subsection{Priming and Working Memory Test result}

To assess the model’s capacity for transient, single-trial overrides of habitual inference, we implemented a one-shot priming experiment with an explicit working memory module. In this paradigm, the model is briefly primed with a high-certainty context (toy definitely moved to the basket), and then immediately probed with an ambiguous context—measuring whether and how quickly belief inference shifts.

\begin{figure}[h]
  \centering
  \includegraphics[width=0.35\textwidth]{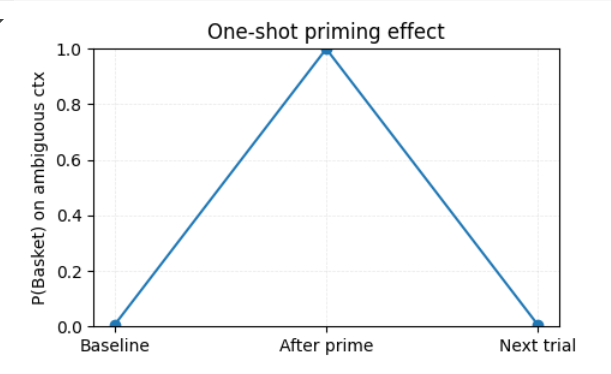}
  \caption{One-shot priming effect: $P(\textrm{Basket})$ on the ambiguous context spikes to $1.00$ immediately after priming, then returns to the baseline value $0.00$ on the next trial, mirroring fast, transient memory-limited priming observed in human cognition.}
  \label{fig:prime-effect}
\end{figure}

\textbf{Outcome:} As shown in Figure~\ref{fig:prime-effect}, the model's probability of ``Basket'' for the ambiguous context is initially low at baseline ($P(\text{Basket}) = 0.00$). Immediately after the priming event, this probability spikes to $1.00$, evidencing a full System~2 override through working memory. On the subsequent ambiguous probe, with no re-priming, the working memory is empty and the probability returns to baseline ($P(\text{Basket}) = 0.00$), confirming a strict one-shot effect.

\textbf{Interpretation:} This sequence demonstrates that the model’s working memory mechanism enables a targeted, single-use belief revision: only the first post-prime inference is affected, while all subsequent identical contexts revert to the habitual result. This sharply transient, context-driven spike closely parallels classical one-shot priming and correction phenomena in human reasoning, providing computational evidence that the architecture supports not only persistent learning but also rapid, ephemeral adjustments.

\subsection{Cognitive Load and System 2 Fatigue Results}

To explore how cognitive resource constraints modulate flexible inference, we introduced a cognitive load input feature and evaluated its effect on System~2 reasoning and gating behavior.

\begin{figure}[h]
  \centering
  \includegraphics[width=0.35\textwidth]{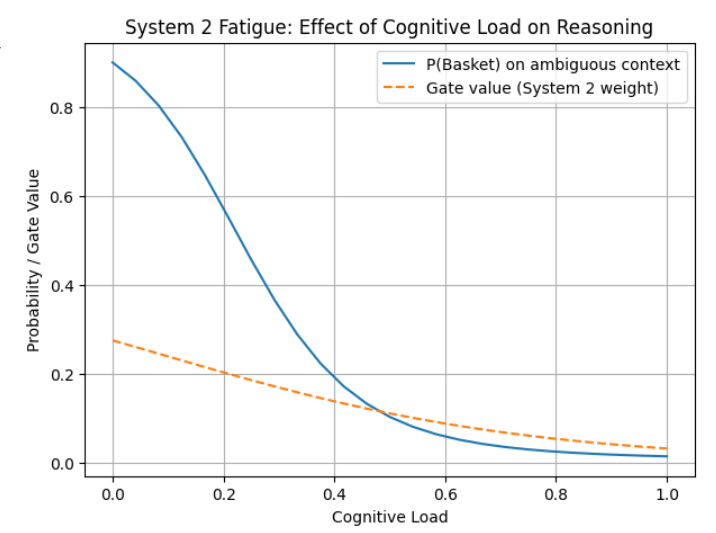}
  \caption{Effect of increasing cognitive load on ambiguous context inference. Both the probability of the correct answer (Basket) and System~2 gate value precipitously decline under higher load, reflecting a transition to System~1 (habitual) dominance.}
  \label{fig:sys2-fatigue}
\end{figure}

\textbf{Ambiguous Reasoning Under Load.}
For an ambiguous context (e.g., Sally returns with uncertain evidence about the toy's location), increasing cognitive load sharply degraded the model's performance. As shown in Figure~\ref{fig:sys2-fatigue}, the probability assigned to ``Basket'' (blended output) drops nonlinearly from $0.89$ (no fatigue) to $0.03$ (maximum fatigue) as the cognitive load increases. Simultaneously, the System~2 gate value decreases from $0.28$ to $0.02$, indicating a shift from controlled override to habitual responding.

\begin{figure}[h]
  \centering
  \includegraphics[width=0.35\textwidth]{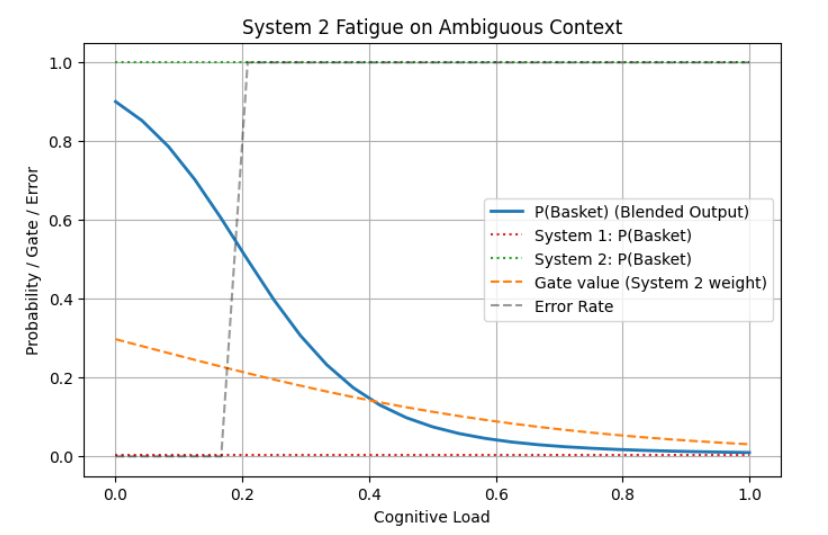}
  \caption{Ambiguous context decomposition: System~2 accuracy remains robust, but the learned gate vanishes under load. The error rate jumps abruptly, aligning with dual-process cognitive theory predictions.}
  \label{fig:fatigue-ambiguous}
\end{figure}

\textbf{System Decomposition and Error Rate.}
Decomposing outputs (Figure~\ref{fig:fatigue-ambiguous}) shows System~2 alone maintains high accuracy ($P(\text{Basket}) \approx 1.0$) regardless of load, while System~1 stays low ($\approx 0.0$). The blended output tracks System~2 at low load, but as gating is suppressed, it shifts abruptly towards System~1, producing a sharp increase in error rate from $0\%$ to $100\%$ at high load.

\begin{figure}[h]
  \centering
  \includegraphics[width=0.35\textwidth]{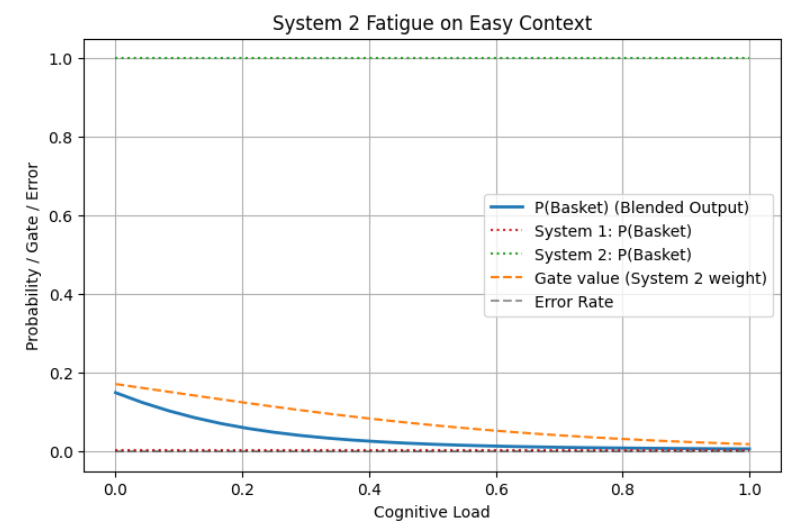}
  \caption{Easy context: Correct inference is robust to cognitive load, since System~1 alone is sufficient and gate suppression has no impact.}
  \label{fig:fatigue-easy}
\end{figure}

\textbf{Easy Context Robustness.}
By contrast, for non-ambiguous, easy contexts (toy definitely in the box), both System~1 and System~2 remain correct. The model's output and error rate are unaffected by increasing cognitive load (Figure~\ref{fig:fatigue-easy}), indicating habitual inference suffices and controlled override is unnecessary.

\textbf{Summary.}
These results quantitatively show that increasing cognitive load selectively impairs controlled System~2 reasoning in ambiguous cases, resulting in rapid performance collapse and a shift to habitual, error-prone inference. The model's learned gating mechanism provides an interpretable measure of reasoning effort, reproducing key predictions of dual-process theory and resource-limited cognition.

\subsection{Framing Effect: Contextual Modulation Without Changing Facts}

To assess the model’s susceptibility to framing effects, we evaluated belief inference under three different narrative framings—negative, neutral, and positive—while holding all factual input features constant (ambiguous scenario: \texttt{[1.0, 0.5, 0.7]}). Framing was encoded as a scalar in the context vector and supplied directly to the gating network.

\begin{figure}[h]
  \centering
  \includegraphics[width=0.35\textwidth]{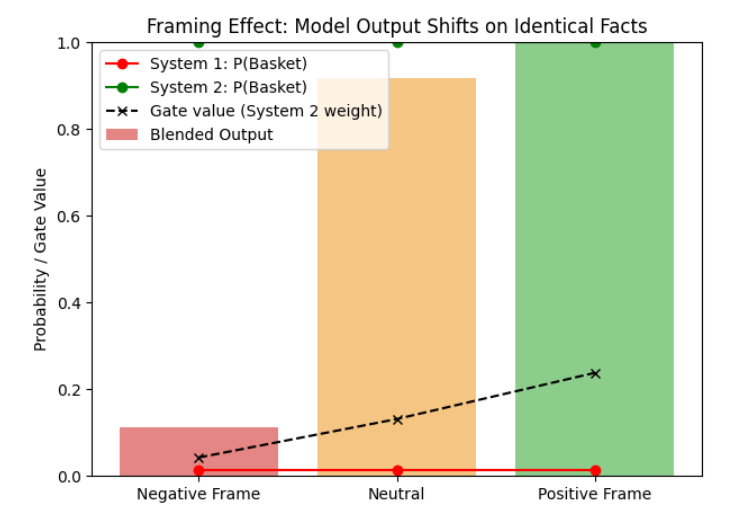}
  \caption{Framing effect: Model output $P(\textrm{Basket})$ and System~2 gate value shift substantially with framing, despite all facts being identical. System~1 remains unaffected; System~2 is recruited more under positive framing.}
  \label{fig:framing-effect}
\end{figure}

\textbf{Outcomes and Quantitative Results.}  
As shown in Figure~\ref{fig:framing-effect}, the blended probability of ``Basket'' assigned by the final model output shifted dramatically as a function of the framing cue, despite identical evidence:
\begin{itemize}
    \item \emph{Negative Frame:} Blended $P(\mathrm{Basket}) = 0.11$, System~1 $= 0.01$, System~2 $= 1.00$, Gate $= 0.04$
    \item \emph{Neutral Frame:} Blended $P(\mathrm{Basket}) = 0.92$, System~1 $= 0.01$, System~2 $= 1.00$, Gate $= 0.13$
    \item \emph{Positive Frame:} Blended $P(\mathrm{Basket}) = 1.00$, System~1 $= 0.01$, System~2 $= 1.00$, Gate $= 0.24$
\end{itemize}

\textbf{Analysis and Interpretation.}  
Both System~1 and System~2 outputs are constant across frames: System~1 always outputs a negligible probability for ``Basket'' ($0.01$), while System~2 strongly prefers ``Basket'' ($1.00$). The key driver of the effect is the learnable gate, which rises sharply from $0.04$ (negative) to $0.24$ (positive), increasing the weight given to System~2 and producing higher certainty in the correct, deliberative inference.

This demonstrates a robust and quantifiable framing bias: superficial narrative cues—without any alteration in factual content—directly modulate the system's internal arbitration, resulting in large, context-driven shifts in final belief. This closely mirrors human cognitive susceptibility to framing effects documented throughout the decision sciences.

These results show that the model is not only sensitive to context in the factual sense, but also vulnerable to purely presentational framing, with system gating mediating when controlled reasoning (System~2) can correct habitual, biased responses. Such findings underline the explanatory power and importance of context-sensitive gating in models of social inference and Theory of Mind.


\section{Discussion and Conclusion}

We present a dual-process neural architecture for Theory of Mind that combines fast, habitual inference with a meta-adaptive, context-sensitive controller modulated by learnable gating. This framework robustly generalizes across novel agents and social contexts, reproducing hallmark human cognitive phenomena—including false-belief reasoning, anchoring bias, priming, cognitive fatigue, and framing effects—within a unified and interpretable model.

Our results highlight the central role of meta-parameter adaptation and gating mechanisms in enabling resource-sensitive, bias-aware inference. Even in simplified environments, the model captures essential aspects of human-like reasoning, offering insights into the computational basis of social cognition. These findings bridge AI and cognitive science, with implications for socially intelligent agents, human-AI collaboration, and the mitigation of reasoning biases.

Future work will extend this approach to more dynamic, real-world multi-agent settings and incorporate richer cognitive and perceptual constraints, moving closer to adaptive, human-aligned machine social reasoning.


\bibliography{aaai2026}

\begin{thebibliography}{24}
\providecommand{\natexlab}[1]{#1}

\bibitem[{Baker et~al.(2017)Baker, Jara-Ettinger, Tenenbaum, and
  Saxe}]{baker2017rational}
Baker, C.~L.; Jara-Ettinger, J.; Tenenbaum, J.~B.; and Saxe, R.~R. 2017.
\newblock Rational quantitative attribution of beliefs, desires and percepts in
  human mentalizing.
\newblock In \emph{Nature Human Behaviour}, volume~1, 0064. Nature Publishing
  Group.

\bibitem[{Baker, Saxe, and Tenenbaum(2011)}]{baker2011bayesian}
Baker, C.~L.; Saxe, R.; and Tenenbaum, J.~B. 2011.
\newblock Bayesian theory of mind: Modeling joint belief‐desire attribution.
\newblock \emph{Proceedings of the Annual Meeting of the Cognitive Science
  Society}, 33(33): 2469--2474.

\bibitem[{Battaglia et~al.(2018)Battaglia, Hamrick, Bapst, Sanchez-Gonzalez,
  Zambaldi, Malinowski, Tacchetti, Raposo, Santoro, Faulkner
  et~al.}]{battaglia2018relational}
Battaglia, P.~W.; Hamrick, J.~B.; Bapst, V.; Sanchez-Gonzalez, A.; Zambaldi,
  V.; Malinowski, M.; Tacchetti, A.; Raposo, D.; Santoro, A.; Faulkner, R.;
  et~al. 2018.
\newblock Relational inductive biases, deep learning, and graph networks.
\newblock \emph{arXiv preprint arXiv:1806.01261}.

\bibitem[{Botvinick et~al.(2020)Botvinick, Wang, Dabney, Munos, Momennejad,
  Lithcap et~al.}]{botvinick2020deep}
Botvinick, M.; Wang, J.~X.; Dabney, W.; Munos, R.; Momennejad, I.; Lithcap,
  Z.~C.; et~al. 2020.
\newblock Deep reinforcement learning and its neuroscientific implications.
\newblock \emph{Neuron}, 107(4): 603--616.

\bibitem[{Evans(2008)}]{evans2008dual}
Evans, J. S.~B. 2008.
\newblock Dual-processing accounts of reasoning, judgment, and social
  cognition.
\newblock \emph{Annu. Rev. Psychol.}, 59(1): 255--278.

\bibitem[{Finn, Abbeel, and Levine(2017)}]{finn2017model}
Finn, C.; Abbeel, P.; and Levine, S. 2017.
\newblock Model-agnostic meta-learning for fast adaptation of deep networks.
\newblock In \emph{International conference on machine learning}, 1126--1135.
  PMLR.

\bibitem[{Grant et~al.(2018)Grant, Finn, Levine, Darrell, and
  Griffiths}]{grant2018recasting}
Grant, E.; Finn, C.; Levine, S.; Darrell, T.; and Griffiths, T.~L. 2018.
\newblock Recasting gradient-based meta-learning as hierarchical bayes.
\newblock In \emph{International Conference on Learning Representations}.

\bibitem[{Hosseinpanah et~al.(2023)Hosseinpanah, Lin, Lane, Hosseini, Chandra,
  and Lago}]{hosseinpanah2023social}
Hosseinpanah, A.; Lin, H.; Lane, N.~D.; Hosseini, S.; Chandra, V.; and Lago, P.
  2023.
\newblock Social reasoning in graph neural networks.
\newblock \emph{Frontiers in Artificial Intelligence}, 6: 1245230.

\bibitem[{Kahneman(2003)}]{kahneman2003maps}
Kahneman, D. 2003.
\newblock Maps of bounded rationality: Psychology for behavioral economics.
\newblock \emph{American economic review}, 93(5): 1449--1475.

\bibitem[{Kahneman(2011)}]{kahneman2011thinking}
Kahneman, D. 2011.
\newblock \emph{Thinking, Fast and Slow}.
\newblock New York, NY: Farrar, Straus and Giroux.
\newblock ISBN 978-0-374-53355-7.

\bibitem[{Kipf(2016)}]{kipf2016semi}
Kipf, T. 2016.
\newblock Semi-Supervised Classification with Graph Convolutional Networks.
\newblock \emph{arXiv preprint arXiv:1609.02907}.

\bibitem[{Kosinski(2023)}]{kosinski2023llmtom}
Kosinski, M. 2023.
\newblock Theory of mind may have spontaneously emerged in large language
  models.
\newblock \emph{arXiv preprint arXiv:2302.02083}, 4: 169.

\bibitem[{Kosinski(2024)}]{kosinski2024evaluating}
Kosinski, M. 2024.
\newblock Evaluating large language models in theory of mind tasks.
\newblock \emph{Proceedings of the National Academy of Sciences}, 121(45):
  e2405460121.

\bibitem[{Lake et~al.(2017)Lake, Ullman, Tenenbaum, and
  Gershman}]{lake2017building}
Lake, B.~M.; Ullman, T.~D.; Tenenbaum, J.~B.; and Gershman, S.~J. 2017.
\newblock Building machines that learn and think like people.
\newblock \emph{Behavioral and brain sciences}, 40: e253.

\bibitem[{Mi et~al.(2023)Mi, Xue, Shi, Zhu, Wang, Ong, and Sun}]{mi2023llmtom}
Mi, F.; Xue, R.; Shi, W.; Zhu, Y.; Wang, Y.; Ong, Y.-S.; and Sun, A. 2023.
\newblock Emergent theory of mind in large language models via prompting.
\newblock In \emph{arXiv preprint arXiv:2302.02083}.

\bibitem[{Milli, Lieder, and Griffiths(2021)}]{milli2021rational}
Milli, S.; Lieder, F.; and Griffiths, T.~L. 2021.
\newblock A rational reinterpretation of dual-process theories.
\newblock \emph{Cognition}, 217: 104881.

\bibitem[{Nichol, Achiam, and Schulman(2018)}]{nichol2018first}
Nichol, A.; Achiam, J.; and Schulman, J. 2018.
\newblock On first-order meta-learning algorithms.
\newblock In \emph{arXiv preprint arXiv:1803.02999}.

\bibitem[{Premack and Woodruff(1978)}]{premack1978chimpanzee}
Premack, D.; and Woodruff, G. 1978.
\newblock Does the chimpanzee have a theory of mind?
\newblock \emph{Behavioral and Brain Sciences}, 1(4): 515--526.

\bibitem[{Rabinowitz et~al.(2018)Rabinowitz, Perbet, Song, Zhang, Eslami, and
  Botvinick}]{rabinowitz2018machine}
Rabinowitz, N.; Perbet, F.; Song, F.; Zhang, C.; Eslami, S.~A.; and Botvinick,
  M. 2018.
\newblock Machine theory of mind.
\newblock In \emph{International conference on machine learning}, 4218--4227.
  PMLR.

\bibitem[{Stanovich and West(2000{\natexlab{a}})}]{stanovich2000advancing}
Stanovich, K.~E.; and West, R.~F. 2000{\natexlab{a}}.
\newblock Advancing the rationality debate.
\newblock \emph{Behavioral and brain sciences}, 23(5): 701--717.

\bibitem[{Stanovich and West(2000{\natexlab{b}})}]{stanovich2000individual}
Stanovich, K.~E.; and West, R.~F. 2000{\natexlab{b}}.
\newblock Individual differences in reasoning: Implications for the rationality
  debate?
\newblock \emph{Behavioral and Brain Sciences}, 23(5): 645--665.

\bibitem[{Swaby et~al.(2024)Swaby, Stewart, Harrold, Willis, and
  Palmer}]{swaby2024machine}
Swaby, L.; Stewart, M.; Harrold, D.; Willis, C.; and Palmer, G. 2024.
\newblock Machine Theory of Mind for Autonomous Cyber-Defence.
\newblock \emph{arXiv preprint arXiv:2412.04367}.

\bibitem[{Tversky and Kahneman(1974)}]{tversky1974judgment}
Tversky, A.; and Kahneman, D. 1974.
\newblock Judgment under Uncertainty: Heuristics and Biases: Biases in
  judgments reveal some heuristics of thinking under uncertainty.
\newblock \emph{science}, 185(4157): 1124--1131.

\bibitem[{Ullman et~al.(2017)Ullman, Spelke, Battaglia, and
  Tenenbaum}]{ullman2017mindgames}
Ullman, T.~D.; Spelke, E.~S.; Battaglia, P.~W.; and Tenenbaum, J.~B. 2017.
\newblock Mind games: Game engines as an architecture for intuitive physics and
  psychology.
\newblock \emph{Trends in Cognitive Sciences}, 21(9): 649--665.

\end{thebibliography}


\appendix
\section*{Appendix}

\subsection{Pseudocode and Algorithmic Framework for the OM2M Architecture}

In this subsection, we provide detailed pseudocode descriptions and explanations of the OM2M (One Model, Two Minds) dual-process Theory of Mind architecture discussed in the main text. These algorithms outline the model components, training, inference, and cognitive-effect experiments.

\paragraph{Algorithm 1: OM2M Model Initialization}  
This algorithm describes the instantiation of the core architecture:
\begin{itemize}
  \item \textbf{System 1 (Fast pathway)}: A Graph Convolutional Network (GCN) that encodes the relational graph and includes a learnable agent meta-vector representing persistent beliefs.
  \item \textbf{System 2 (Slow pathway)}: An MLP meta-adaptive controller that takes System 1's output, flattened parameters, and context features to compute rapid parameter adaptations.
  \item \textbf{Context-Gated Blender}: A gating network that dynamically weights the contributions of System 1 and System 2 based on context features like cognitive load or framing.
\end{itemize}

\paragraph{Algorithm 2: System 1 Pretraining}  
This phase trains the GCN and meta-vector on canonical, unambiguous contexts to capture baseline habitual reasoning before the model learns adaptive control.

\paragraph{Algorithm 3: System 2 and Contextual Gating Training}  
Here the meta-controller and gating network are trained over diverse scenarios (including ambiguous or cognitively demanding contexts). The controller predicts context-dependent parameter updates to adapt System 1, while the gating network learns to arbitrate between Systems 1 and 2 dynamically.

\paragraph{Algorithm 4: OM2M Contextual Inference}  
At test time, System 1 produces an initial belief, which is adapted by System 2 via controller-predicted parameter deltas. The gating network blends these outputs based on the context, producing a final context-sensitive belief distribution.

\paragraph{Algorithm 5: Cognitive Effects Experimental Templates}  
These templates summarize experiments demonstrating key cognitive phenomena in OM2M:  
\begin{itemize}
  \item \textbf{Anchoring}: Persistent belief bias after repeated exposure to a context.
  \item \textbf{Priming}: Transient, one-shot bias induced by salient recent experience stored in working memory.
  \item \textbf{Fatigue}: Modeling decline of System 2 engagement with increasing cognitive load.
  \item \textbf{Framing}: How narrative framing cues modulate System 2 engagement and final outputs with identical factual inputs.
\end{itemize}

\paragraph{Additional Notes}  
The explicit flattening and unflattening of GCN parameters and meta-vector enable fine-grained meta-adaptive updates by the controller. The context vector encodes presence cues, probabilistic information, and cognitive/affective features, facilitating flexible, context-sensitive arbitration akin to human dual-process reasoning.

This modular set of algorithms and their explanations provides a foundation for replicability and conceptual clarity for the OM2M architecture and the cognitive phenomena it models.

\begin{algorithm}[H]
\footnotesize
\caption{OM2M: Context-Gated Dual-Process Social Reasoning}
\begin{algorithmic}[1]
\STATE \textbf{Input:} Relational graph $G=(V,E)$, node features $F_V$, context vector $c$
\STATE Instantiate GCN encoder with parameters $\theta$, agent meta-vector $m$ \hfill // System 1
\STATE Instantiate MLP Controller with parameters $\phi$ \hfill // System 2
\STATE Instantiate context-gated blending network with parameters $\psi$
\STATE \textbf{Output:} OM2M model $(\theta, \phi, \psi, m)$
\end{algorithmic}
\end{algorithm}

\begin{algorithm}[H]
\footnotesize
\caption{System 1 Pretraining (Habit Pathway)}
\begin{algorithmic}[1]
\STATE \textbf{Input:} Canonical ToM contexts, meta-vector $m$
\FOR{each train context $c_{\text{train}}$}
  \STATE $h \gets \mathrm{ReLU}(\mathrm{GCN}(F_V, E))$
  \STATE $h_{\text{agent}} \gets [h[\text{agent}]\, \|\, m]$
  \STATE $\hat{y} \gets \mathrm{metaMLP}(h_{\text{agent}})$
  \STATE Compute loss: $\mathcal{L}_\text{habit} \gets \mathrm{CELoss}(\hat{y}, y_{\text{train}})$
  \STATE Update $\theta, m$ via gradient descent
\ENDFOR
\STATE \textbf{Output:} Pretrained GCN and meta-vector
\end{algorithmic}
\end{algorithm}

\begin{algorithm}[H]
\footnotesize
\caption{System 2 and Contextual Gating Training}
\begin{algorithmic}[1]
\STATE \textbf{Input:} Diverse training set $\{(G_i, F_{V,i}, E_i, c_i, y_i)\}$
\FOR{epoch $=1$ to $T_2$}
  \FOR{each instance $(G, F_V, E, c, y)$}
    \STATE $y_1, h \gets \mathrm{System1}(F_V, E, m)$
    \STATE $\theta_\text{flat} \gets \mathrm{flatten}(\theta, m)$
    \STATE $\Delta\theta \gets \mathrm{Controller}(y_1, \theta_\text{flat}, c;\phi)$
    \STATE $\theta' \gets \theta_\text{flat} + \Delta\theta$
    \STATE $y_2, h' \gets \mathrm{System1}(F_V, E, m;\ \mathrm{params}=\theta')$
    \STATE $g \gets \mathrm{Gate}(c;\psi)$
    \STATE $\hat{y} \gets g\cdot y_2 + (1-g)\cdot y_1$
    \STATE Compute loss $\mathcal{L} \gets \mathrm{CELoss}(\hat{y}, y)$
    \STATE Update $\phi, \psi, m$ via gradient descent
  \ENDFOR
\ENDFOR
\STATE \textbf{Output:} Trained MLP Controller and context-gated network
\end{algorithmic}
\end{algorithm}

\begin{algorithm}[H]
\footnotesize
\caption{OM2M Contextual Inference (Test Phase)}
\begin{algorithmic}[1]
\STATE \textbf{Input:} Test graph $G$, features $F_V$, context $c$
\STATE $y_1 \gets \mathrm{System1}(F_V, E, m)$
\STATE $\theta_\text{flat} \gets \mathrm{flatten}(\theta, m)$
\STATE $\Delta\theta \gets \mathrm{Controller}(y_1, \theta_\text{flat}, c)$
\STATE $\theta' \gets \theta_\text{flat} + \Delta\theta$
\STATE $y_2 \gets \mathrm{System1}(F_V, E, m; \mathrm{params}=\theta')$
\STATE $g \gets \mathrm{Gate}(c)$
\STATE $\hat{y} \gets g\cdot y_2 + (1-g)\cdot y_1$
\STATE \textbf{Return:} $\hat{y}$, $g$
\end{algorithmic}
\end{algorithm}

\begin{algorithm}[H]
\footnotesize
\caption{Cognitive Effects Experimental Templates (OM2M)}
\begin{algorithmic}[1]
\STATE \textbf{Anchoring:} Repeat-train/test on a single context; evaluate belief inertia.
\STATE \textbf{Priming:} Apply one-shot controller/gate from primed context; probe effect on ambiguous inference.
\STATE \textbf{Fatigue:} Vary cognitive load in $c$; record decrease in $g$ and accuracy.
\STATE \textbf{Framing:} Input different frame values for $c$; observe effect on $g$ and outcome.
\end{algorithmic}
\end{algorithm}

\appendix
\makeatletter
\@ifundefined{isChecklistMainFile}{
  \newif\ifreproStandalone
  \reproStandalonetrue
}{
  \newif\ifreproStandalone
  \reproStandalonefalse
}
\makeatother

\ifreproStandalone
\documentclass[letterpaper]{article}
\usepackage[submission]{aaai2026}
\setlength{\pdfpagewidth}{8.5in}
\setlength{\pdfpageheight}{11in}
\usepackage{times}
\usepackage{helvet}
\usepackage{courier}
\usepackage{xcolor}
\frenchspacing

\begin{document}
\fi
\setlength{\leftmargini}{20pt}
\makeatletter\def\@listi{\leftmargin\leftmargini \topsep .5em \parsep .5em \itemsep .5em}
\def\@listii{\leftmargin\leftmarginii \labelwidth\leftmarginii \advance\labelwidth-\labelsep \topsep .4em \parsep .4em \itemsep .4em}
\def\@listiii{\leftmargin\leftmarginiii \labelwidth\leftmarginiii \advance\labelwidth-\labelsep \topsep .4em \parsep .4em \itemsep .4em}\makeatother

\setcounter{secnumdepth}{0}
\renewcommand\thesubsection{\arabic{subsection}}
\renewcommand\labelenumi{\thesubsection.\arabic{enumi}}

\newcounter{checksubsection}
\newcounter{checkitem}[checksubsection]

\newcommand{\checksubsection}[1]{%
  \refstepcounter{checksubsection}%
  \paragraph{\arabic{checksubsection}. #1}%
  \setcounter{checkitem}{0}%
}

\newcommand{\checkitem}{%
  \refstepcounter{checkitem}%
  \item[\arabic{checksubsection}.\arabic{checkitem}.]%
}
\newcommand{\question}[2]{\normalcolor\checkitem #1 #2 \color{blue}}
\newcommand{\ifyespoints}[1]{\makebox[0pt][l]{\hspace{-15pt}\normalcolor #1}}

\end{document}
\fi
\end{document}